%% file: main.tex
\documentclass[letterpaper, 10 pt, conference]{ieeeconf}

\IEEEoverridecommandlockouts
\usepackage{hhline}
\usepackage{colortbl}
\usepackage{algpseudocode}
\usepackage{amsmath}
\usepackage[ruled]{algorithm2e}

\usepackage{graphicx}
\usepackage{amsmath}
\usepackage{amssymb}
\usepackage{hyperref,xcolor}
\usepackage[nocompress]{cite}
\usepackage{float}
\usepackage{aliascnt}
\usepackage{gensymb}
\usepackage{colortbl}
\usepackage{stfloats}
\usepackage{subfigure}
\usepackage{xcolor}
\usepackage{mathtools,amssymb,lipsum}

\usepackage{cuted}
\title{\LARGE \bf
Accurate, Low-Latency Visual Perception for Autonomous Racing:
Challenges, Mechanisms, and Practical Solutions
}

\author{Kieran Strobel$^{1}$, Sibo Zhu$^{1}$, Raphael Chang$^{1}$, and Skanda Koppula$^{2}$
\thanks{$^{1}$ MIT, Cambridge, MA \url{kstrobel,sibozhu,raphc@mit.edu}}%
\thanks{$^{2}$ Google DeepMind, London, UK N1C 4AG \url{skandak@google.com}}%
\thanks{$^{3}$ Open-source repositories/data: \url{driverless.mit.edu/cv-core}}%
\thanks{Work supported by MIT Racing sponsors : \url{driverless.mit.edu}}%
}

\begin{document}
\maketitle
\thispagestyle{empty}
\pagestyle{empty}

\input{00_abstract.tex}
\input{01_introduction.tex}
\input{02_related_work.tex}
\input{03_system_task_and_requirements.tex}
\input{04_hardware_Overview.tex}
\input{05_Software_Stack.tex}
\input{06_Evaluation.tex}
\input{07_Conclusions.tex}

\addtolength{\textheight}{0cm}

\section*{ACKNOWLEDGMENT}
We thank all the members, advisors, and generous sponsors of DUT/MIT Racing for making this project possible.

\clearpage 

\bibliographystyle{IEEEtran}
\bibliography{IEEEabrv,refs}

\end{document}

%% file: 00_abstract.tex
\begin{abstract}

Autonomous racing provides the opportunity to test safety-critical perception pipelines at their limit. This paper describes the practical challenges and solutions to applying state-of-the-art computer vision algorithms to build a low-latency, high-accuracy perception system for DUT18 Driverless (DUT18D), a 4WD electric race car with podium finishes at all Formula Driverless competitions for which it raced. The key components of DUT18D include YOLOv3-based object detection, pose estimation, and time synchronization on its dual stereovision/monovision camera setup. We highlight modifications required to adapt perception CNNs to racing domains, improvements to loss functions used for pose estimation, and methodologies for sub-microsecond camera synchronization among other improvements. We perform a thorough experimental evaluation of the system, demonstrating its accuracy and low-latency in real-world racing scenarios.
\end{abstract}

%% file: 01_introduction.tex
\section{INTRODUCTION}
\vspace{-1mm}

Autonomous racing presents a unique opportunity to test commonly-applied, safety-critical perception and autonomy algorithms in extreme situations at the limit of vehicle handling. As such, work on autonomous race cars has seen a strong uptick since the 2004 DARPA Challenge\textcolor{gray}{\cite{darpa}}, from both industry\textcolor{gray}{\cite{roborace, selfracingcars}} and academia\textcolor{gray}{\cite{amzwholepaper, amzmpc, amzicra1, amzicra2, fsaeteam1, fsaeteam2}}. In recent years, autonomous vehicles have grown increasingly reliant on camera-based perception\textcolor{gray}{\cite{amzvisionthesis, fsaeteam4}}, due to the sensor modality's low cost and high information density. This paper contributes a full-stack design used to translate state-of-the-art computer vision algorithms into an accurate, low-latency computer vision system used on DUT18 Driverless, a 4WD electric racecar with podium finishes at all Formula Driverless competitions in which it raced\textcolor{gray}{\cite{fsg, fsitaly_results, fsg_results}}.

\begin{figure}[h]
\centering
\includegraphics[width=0.47\textwidth]{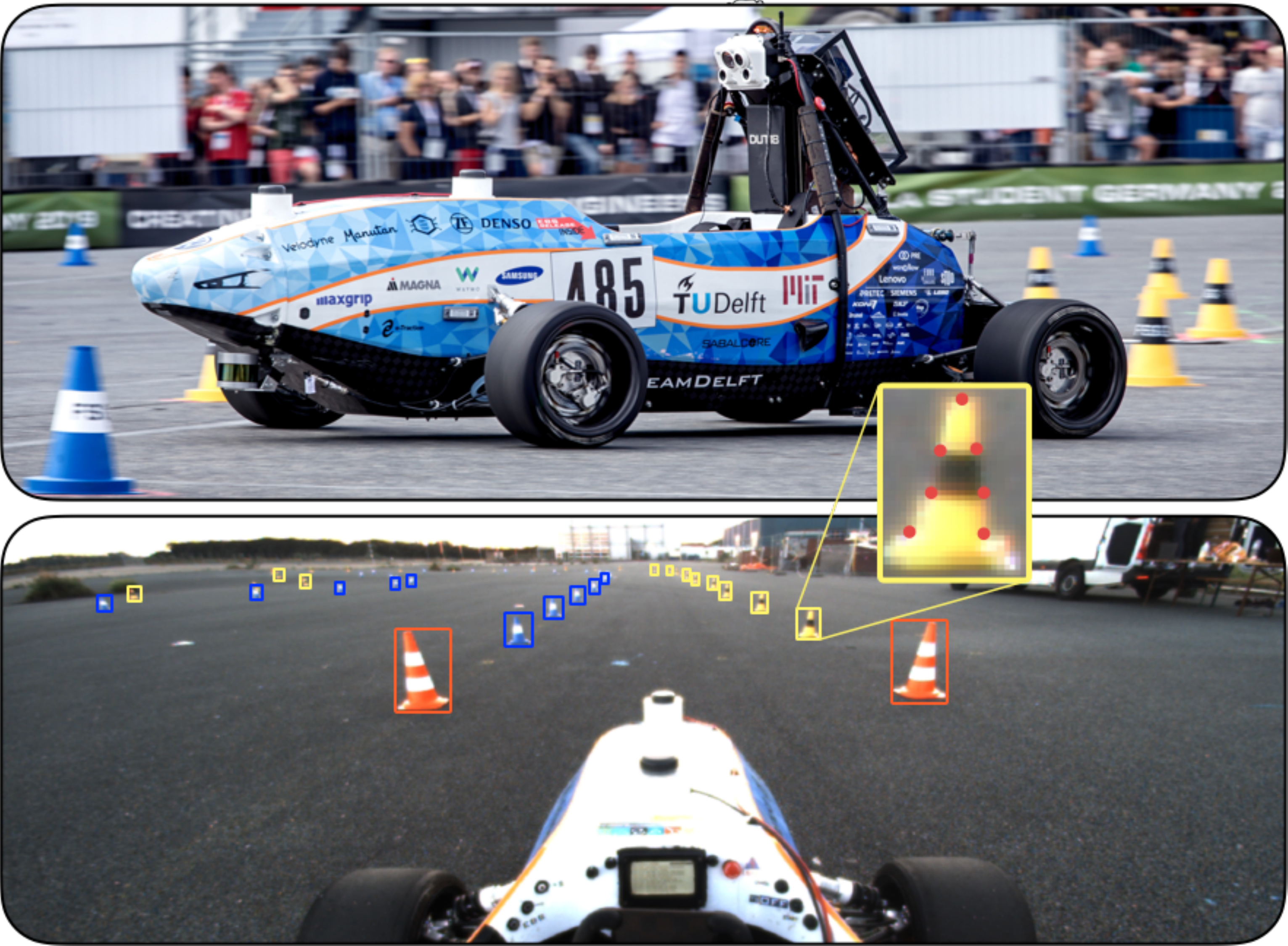}
\caption{DUT18D racing on track, with vision hardware mounted on the car's main roll hoop (top). Frame captured from the monocular camera (bottom), with an example detected track landmark (inset)}
\vspace{-7mm}
\label{fig:dut18d}
\end{figure}

Of critical importance in autonomous driving is the latency of the hardware and software stack. Latency directly impacts the safety and responsiveness of autonomous vehicles. Based on measurements from DUT18D, the visual perception system in an autonomous vehicle dominates the latency of the entire autonomy stack (perception, mapping, state estimation, planning, and controls), occupying nearly 60\%\textcolor{gray}{\cite{asf}} of the end-to-end latency. Prior work corroborates this observation citing the intense computational load of modern visual processing as the source of latency bottlenecks in high-speed autonomous vehicles\textcolor{gray}{\cite{perceptionlatency}}. Despite its importance, low-latency visual perception remains riddled with practical challenges across the entire stack, from noisy image capture and data transmission to accurate positioning in an unmapped environment. To our knowledge, there remains no available prior work detailing the full-stack design of a high-accuracy, low-latency perception system for autonomous driving.

To this end, we describe the design and evaluation of a camera-based perception system employed by DUT18D -- an autonomous racecar boasting 1.2g static accelerations and 100kph top speeds built for the Formula Student Driverless competitions\textcolor{gray}{\cite{fsg_rules}}. This camera-based perception system was designed to perceive and position landmarks on a map using multiple CNNs for object detection and depth-estimation. The design targeted high-accuracy and low-latency across outdoor weather and lighting conditions in the USA, Italy, Germany, and the Netherlands. The car is also equipped with an isolated LiDAR system to improve safety through sensor redundancy, however this will not be the focus of this work.

This paper provides four key contributions: (i) an open design of a thoroughly-tested low-latency vision stack for high-performance autonomous racing, (ii) a comprehensive description of our solutions to common bottlenecks deploying state-of-the-art CV algorithms: new techniques for domain adaptation of pre-trained CNN-based object detectors, useful loss function modifications for landmark pose estimation, and microsecond time synchronization of multiple cameras,
(iii) open-source C++ modules for mobile-GPU accelerated DNN inference, landmark pose estimation, and a plug-and-play visual perception system for Formula Student racecars$^{3}$, and (iv) a publicly available 10K+ pose-estimation/bounding-box dataset for traffic cones of multiple colors and sizes$^{3}$. Although developed in the context of Formula Driverless, the methods presented in this work can be adapted to a wide range of perception systems used for autonomous platforms.

%% file: 02_related_work.tex
\section{RELATED WORK}
\vspace{-1mm}

Prior literature in high-performance autonomy has largely focused on controls\textcolor{gray}{\cite{amzmpc, fsaeteam1, fsaeteam4, deeppicar}}, planning\textcolor{gray}{\cite{fsaeteam2, fsaeteam3}}, full-vehicle integration\textcolor{gray}{\cite{roborace, amzicra1}}, or multi-sensor fusion\textcolor{gray}{\cite{amzicra2}}.  This paper extends these works to the perception domain, solving challenges pertaining to full-stack visual perception system.

Other prior works related to computer vision tasks have explored more narrow solutions to monocular depth estimation\textcolor{gray}{\cite{monocular1, monocular2, monocular3, monocular4}}, stereo depth and pose estimation\textcolor{gray}{\cite{stereo1, stereo2, stereo3, stereo4}}, fused monocular/stereo pose estimation\textcolor{gray}{\cite{dhall2019real}}, 2D object detection\textcolor{gray}{\cite{yolo, squeeze_net, mobilenet, fasterrcnn, realtimetracking}}, obstacle detection\textcolor{gray}{\cite{drone1, drone2, drone3}}, and instance segmentation\textcolor{gray}{\cite{sun2020see, semanticsegmentation, joint_detection2, joint_detection1}} among others\textcolor{gray}{\cite{othercv1, othercv2, othercv3}}. These papers are useful to guide system design, but ignore critical system level challenges that arise in real world systems: training dataset domain mismatch, lens effects, timing issues that impact the latency/accuracy trade-off, and more. To the best of our knowledge, this work is the first to (1) tie together a system that addresses these emergent issues, (2) thoroughly details the solutions and methodology that went into designing a full perception system, and (3) validates system performance using fixed dataset benchmarks and competition environments.

%% file: 03_system_task_and_requirements.tex
\section{SYSTEM TASK AND REQUIREMENTS}
\vspace{-1mm}

\begin{figure}[h]
\vspace{-2mm}
\centering
\includegraphics[width=0.45\textwidth]{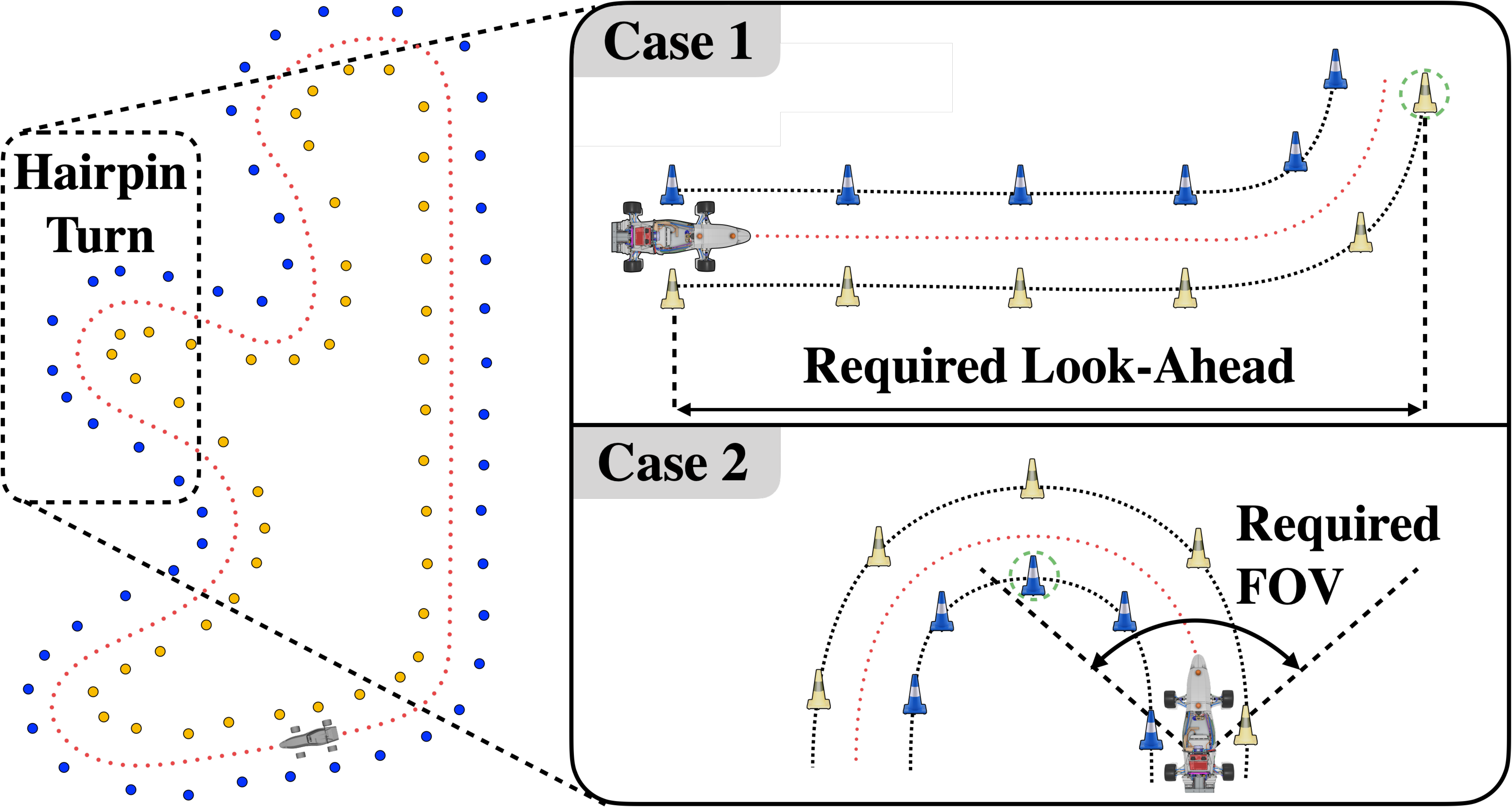}
\vspace{-2mm}
\caption{Sample autocross track (left). Cases that drive minimum look-ahead requirements (top right) and FOV requirements (bottom right).}
\vspace{-2mm}
\label{fig:autoX}
\end{figure}

The goal of this perception system is to accurately localize environment landmarks (traffic cones for this application) that demarcate the racetrack while adhering to regulations set by Formula Student Driverless\textcolor{gray}{\cite{fsg_rules}}. The track is delineated by blue cones on the left, yellow cones on the right, and orange cones at the start and finish. The downstream mapping system then uses these landmarks to create and update the track map with a sample illustrated in Figure~\textcolor{red}{\ref{fig:autoX}}.

To ensure the overall vehicle performance is never bottle-necked by the perception system we identify four system requirements quantifying efficiency, accuracy, and coverage area. With minor adjustments these requirements and the process used to obtain values for them can be generalized to assist in the design of other visual perception systems:
\begin{enumerate}
    \item \emph{Latency}: total time from a landmark coming into the perception system's view to the time it is localized. This puts a requirement on the system's 'efficiency'.
    \item \emph{Mapping Accuracy}: max allowed landmark localization error. This puts a requirement on system accuracy.
    \item \emph{Horizontal Field-of-View (FOV)}: the arc of visibility.
    \item \emph{Look-ahead Distance}: longest straight-line distance in which accuracy is maintained. With FOV, these put requirements on the system's coverage area.
\end{enumerate}

To guide design choices we derived quantitative values for each requirement. \emph{Mapping accuracy} was driven by the downstream mapper's data association capabilities\textcolor{gray}{\cite{fastslam1, fastslam2}} dictating a maximum localization error of 0.5m. For \emph{latency}, we developed a kinematic model of DUT18D and simulated a safe, emergency stop from top speed (25m/s) during a Formula Student acceleration run. From this, we derived a maximum 350ms view-to-actuation latency of which downstream systems required 150ms translating to a perception latency budget of 200ms --  consistent with prior work\textcolor{gray}{\cite{etsi2013101}}. \emph{Horizontal FOV} is lower-bounded by unmapped hairpin U-turns, where the competition rules dictate the minimum radius. As shown in Figure\textcolor{red}{~\ref{fig:autoX}}, the system must perceive landmarks on the inside apex of the turn in order to plan an optimal trajectory which results in a minimum FOV of 101\degree. Required \emph{look-ahead distance} depends on the full-stack-latency and vehicle deceleration rates. To quantify this we built a kinematic model of the vehicle at top track-drive speed (15 m/s) going into a hairpin turn  as shown in Figure~\textcolor{red}{\ref{fig:autoX}}. This yields the largest braking distance and, in turn, the maximum required look-ahead distance, 19.6m.

%% file: 04_hardware_Overview.tex
\section{HARDWARE STACK}

\begin{figure}[b!]
    \vspace{-6mm}
    \centering
    \includegraphics[width=0.47\textwidth]{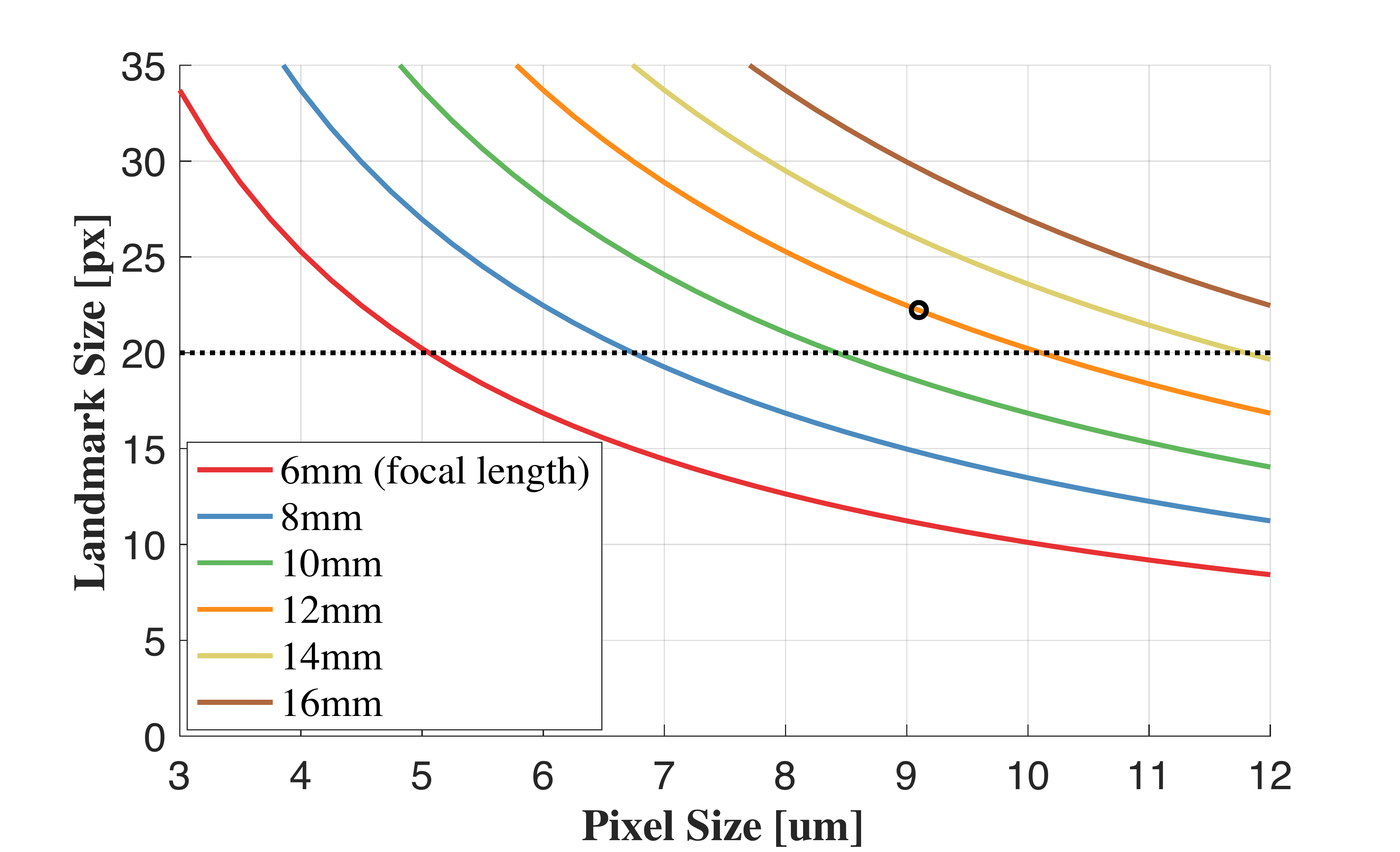}
    \vspace{-3mm}
    \caption{Long-range camera options. The chosen configuration (4.55um 2x binned pixels/12mm focal length) is plotted for reference.}
    \label{fig:look-ahead}
    \vspace{-1mm}
\end{figure}

A difficulty which arises during the design process is balancing the trade offs between software and hardware performance. A small focal-length lens, for example, will provide a large field-of-view but result in few pixels on the long-range landmarks for the algorithms to localize. This can be compensated for by using sensors with small physical pixels but will then result in dark images due to the small light capture area. To select an appropriate hardware configuration, namely lens focal-length and physical camera sensor pixel size, we developed a model based on a pin-hole camera detecting landmarks with known dimensions (0.33m) at the maximum look-ahead distance. As shown in Figure~\textcolor{red}{\ref{fig:look-ahead}}, this can be used to determine the needed software performance, measured by the minimum number of pixels required to accurately localize a landmark, for a given hardware configuration. Consistent with state-of-art work\textcolor{gray}{\cite{yolo,nyu_depth}} we assume 20 pixels are needed, as represented by the dashed line in this figure. The optimal configuration is a system that (1) stays above this line, (2) maximizes pixel size (i.e. maximizing light capture), and (3) minimizes focal length (i.e. maximizing FOV). Figure~\textcolor{red}{\ref{fig:look-ahead}} illustrates how these are mutually exclusive with each other.

\begin{figure}[h]
\vspace{-4mm}
\centering
\includegraphics[width=0.475\textwidth]{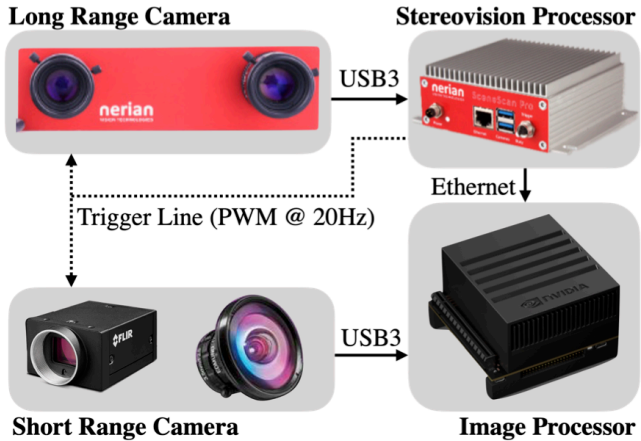}
\vspace{-3mm}
\caption{A 10cm baseline stereo pair made from two Basler daA1600-60um + 12mm lenses (with 2x pixel binning on sensor) were used along with a Pointgrey GS3-U3-23S6C-C + 3.5mm lens. Stereo matching was done on a Nerian FPGA creating low latency dense disparity maps using a semi-global matching algorithm. The disparity map and images from both pipelines were then processed on an Nvidia Jetson Xavier where landmarks are detected and depth estimates are extracted. The localized landmarks were then sent to a separate compute unit for the remainder of the autonomous pipeline.}
\vspace{-3mm}
\label{harware}
\end{figure}

The ideal camera hardware for each of the cases outlined in Figure~\textcolor{red}{\ref{fig:autoX}} have opposing specifications due to the inherent look-ahead distance and FOV trade-off in camera-based systems. To satisfy the requirements for each of these cases a two camera architecture was deployed with a stereo camera used for long-range detection and a monocular camera for short-range detection. An overview of the hardware used for this system are shown in Figure~\textcolor{red}{\ref{harware}}. As will be discussed, our work found that the largest performance benefits were derived from software improvements. In order to focus efforts on software performance, the hardware was selected based on a configuration which satisfied the requirements, rather than determining a globally optimum layout.

\begin{figure*}[b!]
\vspace{-3mm}
\centering
\includegraphics[width=0.999\textwidth]{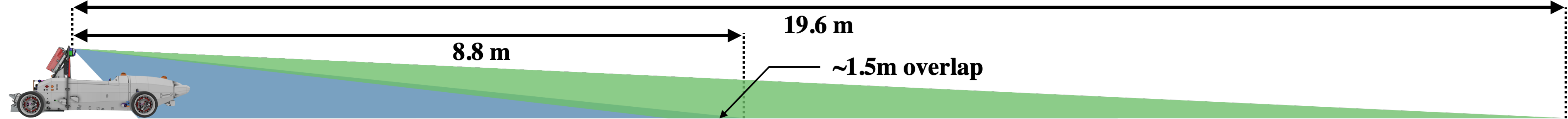}
\vspace{-6mm}
\caption{Camera Viewing Regions. For occlusion mitigation the cameras were mounted high within the vehicle roll hoop. Processing/transfer latency was improved by cropping images on sensor to remove pixels above the horizon. The green and blue regions show the long and short range regions, respectively.}
\label{bird_eye_camera}
\end{figure*}

Ideally, a dual stereovision system would be used to reduce depth estimation complexity; this was infeasible due to packaging constraints by the competition rules. Benefits, however, of monocular vision include reduced part count and calibration efforts which proved to be advantageous. To simplify the more complex monocular depth estimation process, the monocular camera was used for short-range detection rather than long-range detection as the pixel space location of a landmark on a globally flat surface is more sensitive to pose when up-close, making it easier to localize. The viewing regions of our system are shown in Figure~\textcolor{red}{\ref{bird_eye_camera}}.

%% file: 05_Software_Stack.tex
\section{SOFTWARE STACK}
\vspace{-1mm}

We carefully designed the software stack to complement the chosen hardware given that performance is a strong function of the software architecture. To ensure low latency landmark localization our software stack follows three main steps to produce pose estimates from raw images:
\begin{enumerate}
    \item Data Acquisition: Image streams are captured, disparity matched (for the stereovision pipeline), synchronized, and transferred to the Jetson Xavier.
    \item 2D Space Localization: Using a NN-based approach with a modified YoloV3 network\textcolor{gray}{\cite{yolo}}, landmarks are detected and outlined by bounding boxes in the images.
    \item 3D Space Localization: Depth from each landmark is extracted by a clustering-based approach for stereovision and a NN-based approach for monocular vision.
\end{enumerate}

\subsection{Data Acquisition}

\begin{figure*}[t!]
\centering
\includegraphics[width=0.82\textwidth]{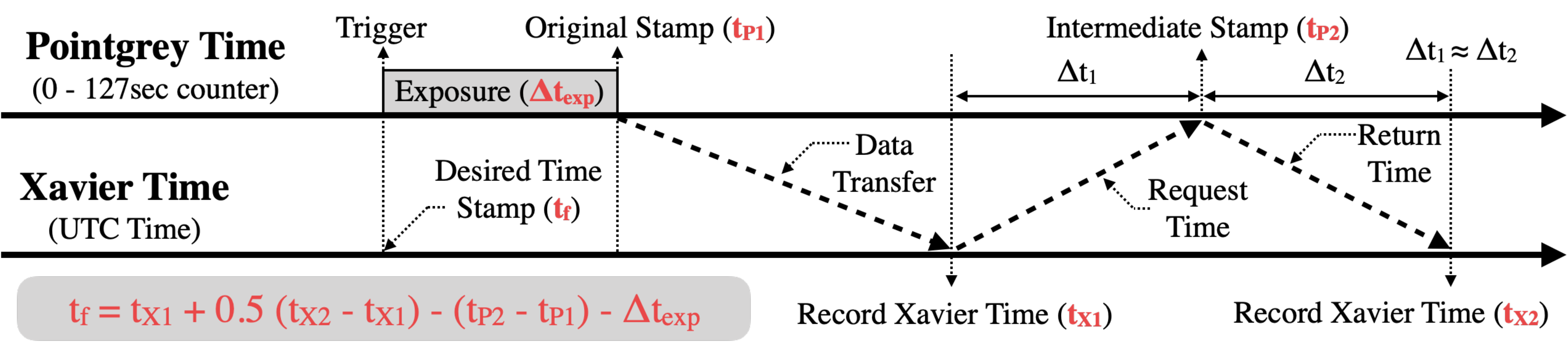}
\vspace{-2mm}
\caption{Monocular camera time synchronization. Variables used for time-translation calculations are highlighted in red for clarity. As the monocular camera is triggered off of the same signal as the stereovision system, the timestamp ($t_f$) calculated from this process is then re-stamped with the closest stereovision image timestamp. If this restamping process changes the timestamp by more than 15ms then the frame is thrown out.}
\vspace{-6mm}
\label{mono}
\end{figure*}

Without precise time stamps accompanying each image, inaccuracies as small as 10ms (i.e. the typical USB3 transport latency for HD images) can result in 25cm localization errors at high speeds -- 50\% of the allowable localization error. To solve this, we developed a system where both pipelines are triggered by a hardware timestamped signal generated by the FPGA resulting in sub-microsecond accuracy, translating to sub-millimeter localization errors contributed by the time synchronization. The stereo-pair images are given this timestamp which is synchronized to the Xavier’s 
master clock using PTP. To ensure this same level of accuracy for the USB3 monocular camera, a custom time-translation process shown in Figure~\textcolor{red}{\ref{mono}} was developed resulting in sub-microsecond synchronization over USB3.

In the case of the monocular vision pipeline, there is an inherent trade-off between localization accuracy and latency. Methods for monocular depth estimation outlined in Section II improve with increasing resolution, whereas the latency of CNN based object detection algorithms scale linearly with resolution. To side-step this issue images from the monocular camera are captured at 1600x640 and transferred to the Jetson Xavier where they are later software binned to 800x320. The low-resolution images are used for object detection and the high-resolution images are used for depth estimation. This allows for high accuracy depth estimation while maintaining low-latency object detection. As the system performance is bottlenecked by the accuracy of the monocular depth estimation process, the use of low-resolution images for object detection does not result in a performance reduction. For the stereovision pipeline images are also captured at 1600x640, but are 2x hardware binned to increase the amount of light captured by each pixel. The performance of the stereo matching process met all requirements without the need for higher resolution.

\subsection{2D Localization: YoloV3}

\begin{figure}[h]
\vspace{-2mm}
\centering
\includegraphics[width=0.44\textwidth]{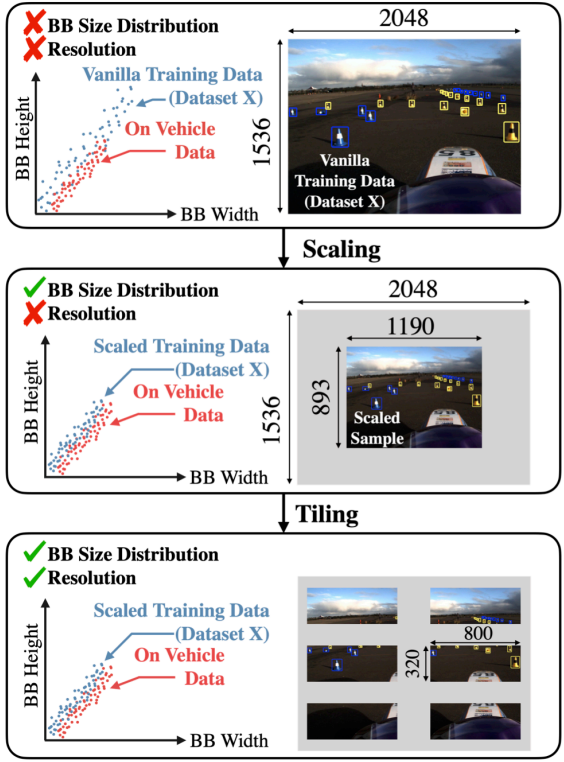}
\caption{CV-C Dataloader Pre-processing Stages. Each set of training images from a specific sensor/lens/perspective combination are uniformly rescaled such that their landmark size distributions match that of the camera system on vehicle. Each training image is then padded if smaller than 800x320 or split up into multiple images if larger than 800x320.}
\vspace{-2mm}
\label{data pre-processing}
\end{figure}

A critical challenge for NN-based perception systems is poor generalization across different domains and perspectives. To reduce this effect on our models, training data was collected on multiple cameras and lenses from various perspectives in different settings. A drawback of this process is the training set distribution of landmark bounding box sizes (in pixels) no longer represents what would be seen by the network on vehicle. To solve this, we used a process of scaling and tiling each set of images from a specific sensor/lens/perspective combination such that their landmark size distributions was representative of the distribution seen on vehicle. This process is illustrated in Figure~\textcolor{red}{\ref{data pre-processing}}. Since YOLOv3 makes detection width and height predictions by resizing predefined anchor boxes, a k-means clustering algorithm was then run on the post-scaled training data in height and width space to give the network strong priors.

Each bounding box prediction estimates x, y, confidence, width, height, and class (foreground/background) which are penalized during training using loss constants as follows:

\vspace{-4mm}
\begin{equation}
\centering
    \begin{split}
    L_{total}= \gamma_{cls}L_{cls-ce} +\gamma_{BG}L_{BG-bce} +\gamma_{FG}L_{FG-bce} \\
    +\gamma_{xy}(L_{x-mse}+L_{y-mse}) +\gamma_{wh}(L_{w-mse}+L_{h-mse})
    \end{split}
\end{equation}
\vspace{-3mm}

\begin{figure}[h]
\centering
\includegraphics[width=0.49\textwidth]{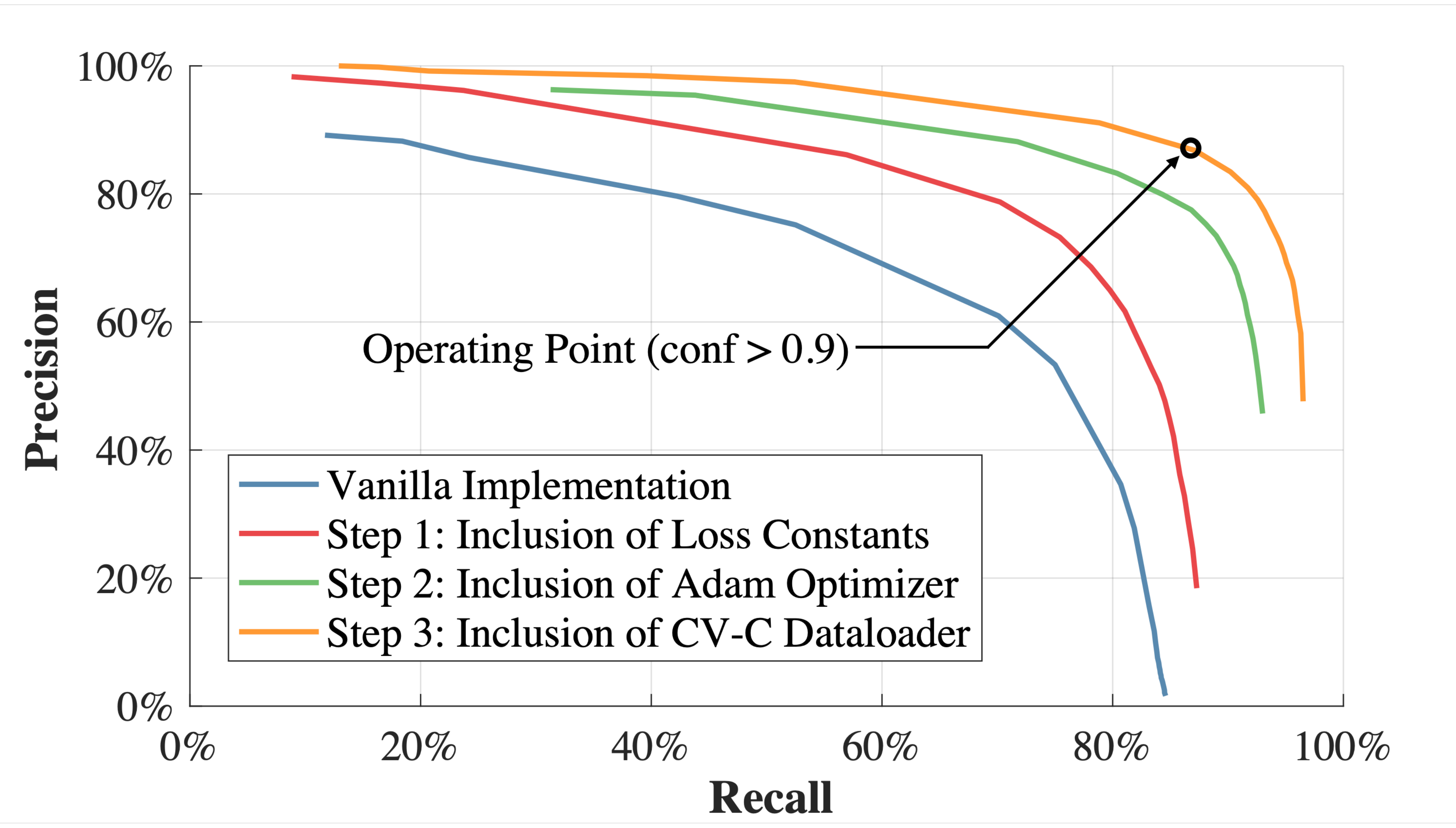}
\vspace{-5mm}
\caption{The compounding benefits for each of the YoloV3 modifications are shown as precision-recall curves with the converged upon values for the loss constants from the optimization process being $\gamma_{BG}$=25.41, $\gamma_{FG}$=0.09, $\gamma_{XY}$=1.92 and $\gamma_{WH}$=1.33. The resulting final mAP was 85.1\% with 87.2\% recall and 86.8\% precision through training on a dataset of 73k landmarks.}
\vspace{-2mm}
\label{yolo_mod}
\end{figure}

To further improve detection accuracy a distributed Bayesian optimization study was run to determine optimal values for these five loss constants. Further gains were obtained by switching from the SGD optimizer in the initial implementation to Adam\textcolor{gray}{\cite{Adam}}. The accumulative performance gains of these modifications are shown in Figure~\textcolor{red}{\ref{yolo_mod}}.

\vspace{-3mm}

To reduce latency, NVIDIA's TensorRT API was used to infer on the batched images using the YoloV3 NN. Further reductions were obtained by calibrating the network weights to int-8 precision, resulting in a 10x latency improvement when compared to a similar PyTorch implementation. Occlusions were found to be problematic for depth estimation downstream. To filter out these overlapping landmarks, aggressive non-maximal suppression thresholds were used.

\subsection{3D Localization: Monocular Vision}
\begin{figure*}[t!]
\centering
\includegraphics[width=0.83\textwidth]{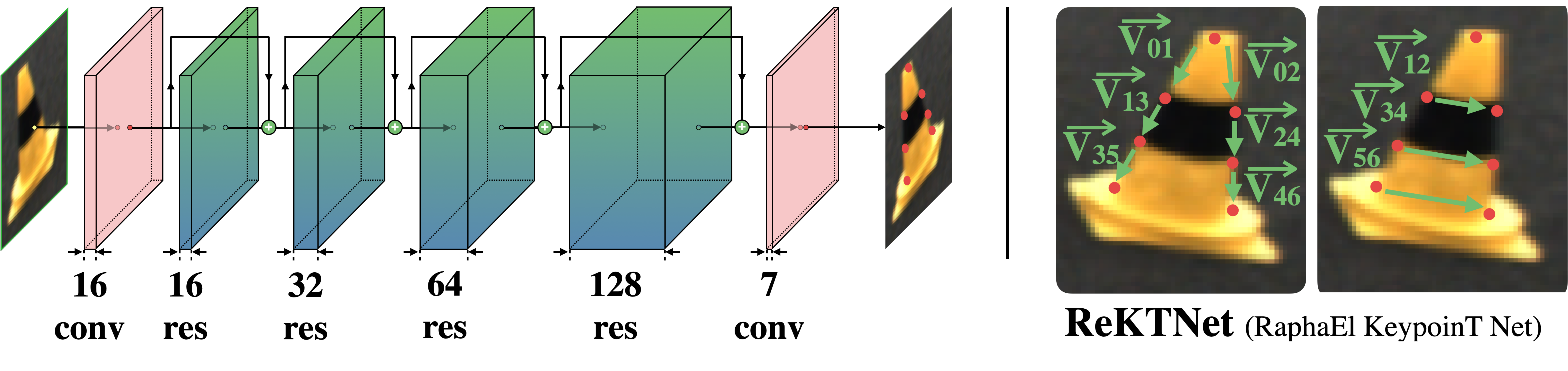}
\vspace{-3.5mm}
\caption{The ReKTNet architecture consists of fully convolutional layers to detect seven keypoints on each cone (left). The dot product between collinear vectors connecting keypoints on a cone is leveraged in the ReKTNet loss function (right).}
\vspace{-6mm}
\label{rekt_loss}
\end{figure*}

Landmark localization accuracy for single camera systems can be improved by using a priori knowledge of landmark dimensions to provide scale to images. We leverage this through the use of an additional residual NN, ReKTNet, trained using 3.2k images, similar to \textcolor{gray}{\cite{amzvisionthesis}}. ReKTNet detects seven keypoints on each YoloV3 detection which are then used in a Perspective-n-Point (PnP) algorithm to get scale and, in turn, depth. The NN architecture and a sample output are shown in Figure~\textcolor{red}{\ref{rekt_loss}}. To ensure the algorithm is made robust to single keypoint outliers, all subset permutations of the keypoints with one point removed are calculated if the reprojection error from the seven keypoint estimate is above a threshold. The lowest error permutation is then used.

To improve accuracy ReKTNet, contains two important network modifications that build on the original implementation which can be generalized to other applications. First, the fully connected output layer was replaced with a convolutional layer to predict a probability heatmap over the image for each keypoint. Convolutional layers outperform FC layers at predicting features which are spatially interrelated while also improving training convergence by having fewer parameters. The expected value over the heatmap\textcolor{gray}{\cite{dsnt}} is then used as the keypoint location. The second modification is an additional term in the loss function to leverage the geometric relationship between keypoints as shown in Figure~\textcolor{red}{\ref{rekt_loss}}. Since the keypoints on the sides of a cone are collinear, the unit vectors between points on these lines must have a unity dot product. One minus the values of these dot products are used directly in the loss function; the same is done for the three horizontal vectors across the cone. As the keypoint locations need to be backpropagated, the differentiable expected value function\textcolor{gray}{\cite{dsnt}} is used to extract heatmap coordinates. Although applied to cones here, this insight can be used in any perception system detecting objects with known dimensions. The final loss function is as follows:

\vspace{-4mm}
\begin{equation}
\begin{aligned}
    \begin{split}
    L_{total}=L_{mse} + \gamma _{horz}(2 - V_{12} \cdot V_{34} - V_{34} \cdot V_{56}) +  \\
    \gamma _{vert}(4-V_{01}\cdot V_{13}- V_{13}\cdot V_{35} - V_{02} \cdot V_{24} - V_{24} \cdot V_{46})  
    \end{split}
\end{aligned}
\end{equation}
\vspace{-5mm}

\begin{figure*}[b!]
\vspace{-2mm}
\centering
\includegraphics[width=0.97\textwidth]{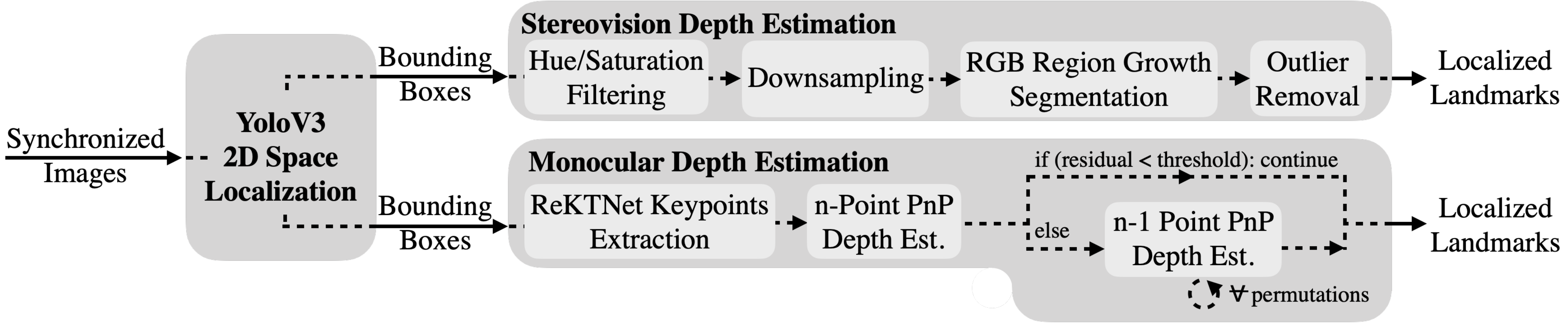}
\vspace{-3mm}
\caption{Depth Estimation Pipeline. After 2D space localization each set of bounding boxes follow a separate process resulting in long-range detection from the stereovision pipeline and short-range detection from the monocular pipeline.}
\label{depth}
\end{figure*}

Similar to our YoloV3 work, a Bayesian optimization framework was used to determine values for the loss constants, resulting in $\gamma_{vert}=0.038$ and $\gamma_{horz}=0.055$. The overall process, along with the stereovision depth estimation process described next, is illustrated in Figure~\textcolor{red}{\ref{depth}}.

\vspace{-1mm}
\subsection{3D Localization: Stereo Vision}
A major challenge in producing stereovision depth estimates is point cloud segmentation of the landmark of interest, especially when there is a lack of texture which results in poor SGM performance. To solve this, we first convert each detection from RGB to HSV and filter out pixels with V or S values below 0.3. This both helps segmentation by removing the asphalt background as well as removes the stripe on each cone as this region lacks texture. To reduce latency, if more than 200 points remain within the bounding box the point cloud is downsampled and then clustered in XYZ and RGB space. The XYZ-centroid of the cluster with the closest average hue to the cone color is used as the cone location. 

\vspace{-4mm}
\begin{figure}[h]
\centering
\includegraphics[width=0.51\textwidth]{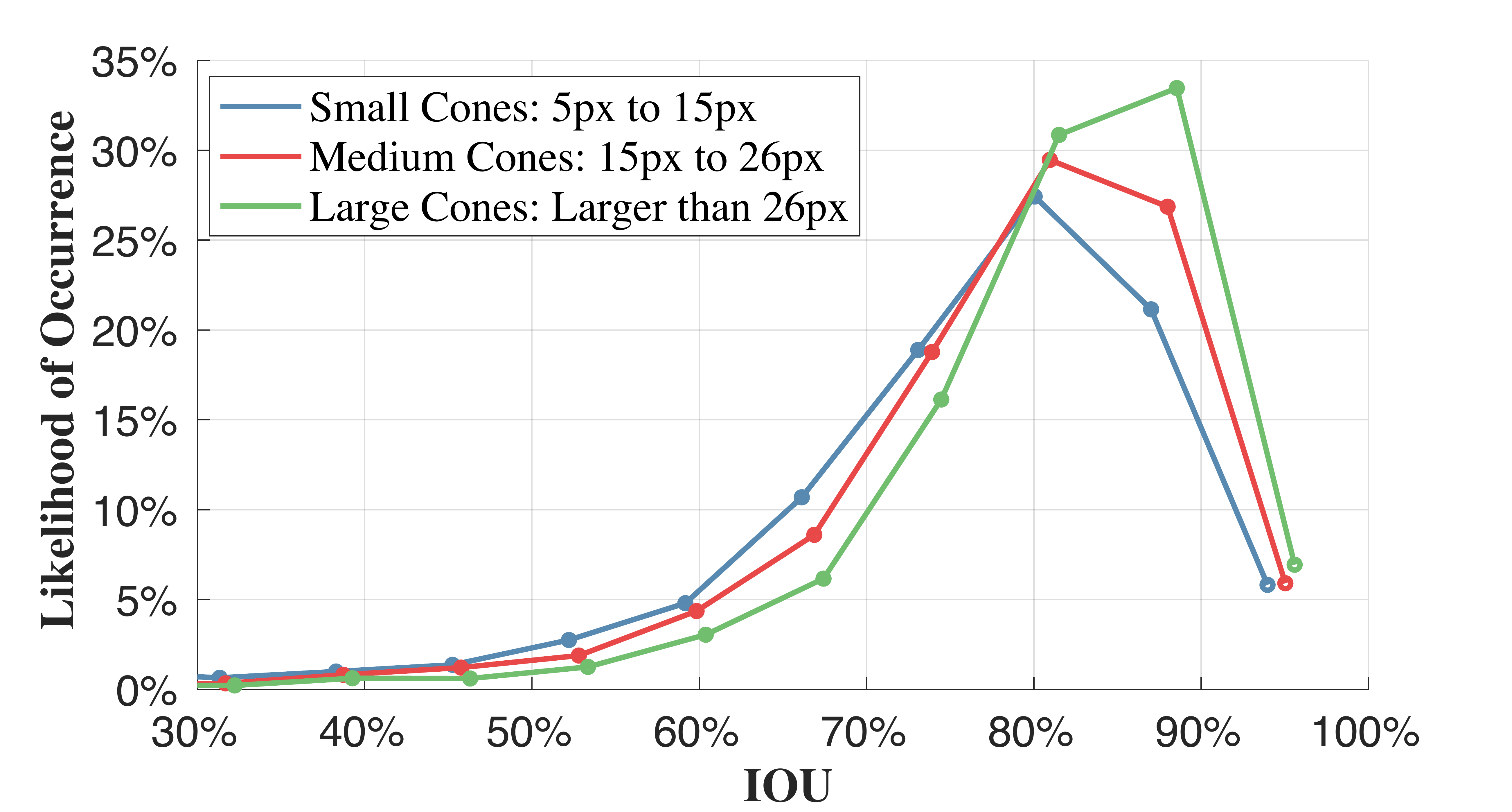}
\vspace{-6mm}
\caption{Landmark detection IOU distribution for various bounding box sizes. Our system shows only minor variations in detection accuracy as a function of landmark size and produces high IOUs with up to 75\% fewer pixels than the 20 pixel requirement of Section III. For mAP calculations, IOUs below 50\% were discarded. 10 bins were used to create the PDF.}
\vspace{-3mm}
\label{fig:IOU}
\end{figure}

%% file: 06_Evaluation.tex
\section{Validation}
\vspace{-1mm}

System validation efforts were focused on validating the four high-level requirements of Section III and the YoloV3 performance. The FOV requirement was satisfied by lens and sensor selection, resulting in a horizontal FOV of 103$^{\circ}$.

To characterize mapping accuracy we first examined 2D detection accuracy across three landmark sizes with results shown in Figure~\textcolor{red}{\ref{fig:IOU}}. Each curve contains $>$24,000 landmarks and was evaluated using a widely-used metric, intersection-over-union (IoU), which measures alignment of our bounding box with ground truth\textcolor{gray}{\cite{yolo}}. We achieve a median IoU of 88\% for large cones, and 83-84\% for smaller cones with an overall mAP of 85.1\% -- 9\% above the current state-of-the-art\textcolor{gray}{\cite{amzvisionthesis}} which enables the mapping accuracy described next.

\begin{figure}[h]
\vspace{-3.5mm}
\centering
\includegraphics[width=0.47\textwidth]{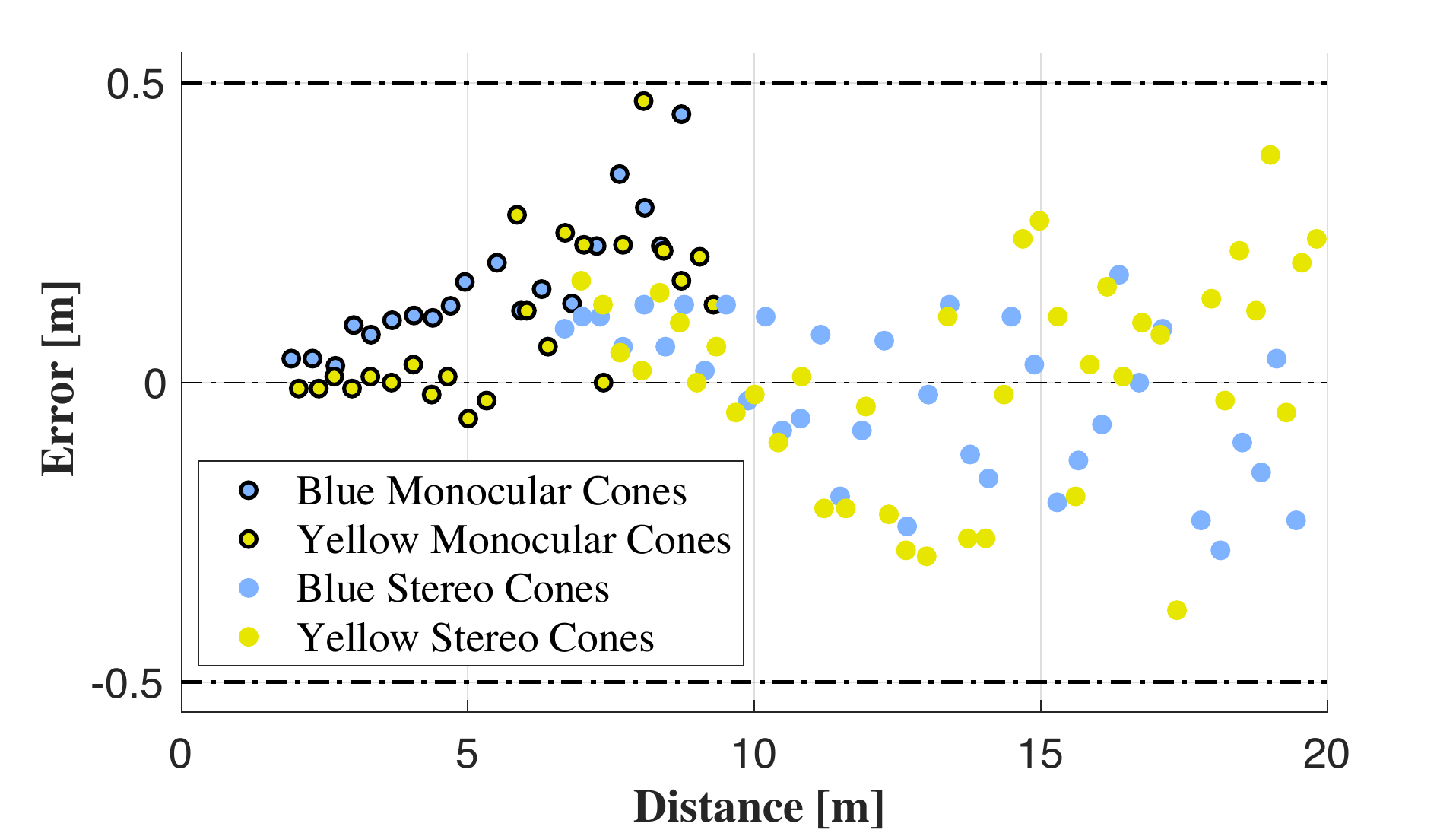}
\vspace{-4mm}
\caption{Depth estimation error vs. distance. Dotted lines provide bounds set by the system requirements. Mean errors for the 100 frames of recorded data were below 0.5m for both pipelines, and standard deviations were below 5cm for the monocular pipeline and 10cm for the stereo pipeline.}
\vspace{-3mm}
\label{depth_error}
\end{figure}

Mapping accuracy and look-ahead were validated in tandem by using the track illustrated in Figure~\textcolor{red}{\ref{fig:autoX}}. The vehicle was statically placed at five different locations and 100 frames were collected to estimate distances of the landmarks in the system's FOV. Ground truth values for Euclidean distances from the cameras were measured using a Leica laser rangefinder to produce the results shown in Figure~\textcolor{red}{\ref{depth_error}}. The experiment was done statically to remove the dependency on the vehicle's state estimation system.

The stereovision pipeline results show a maximum effective disparity error of 0.15 pixels -- a hardware agnostic metric which is 40\% lower than the expected value\textcolor{gray}{\cite{nerian}}. To provide a hardware agnostic comparison for the monocular vision algorithm's performance to the current state-of-the-art\textcolor{gray}{\cite{amzvisionthesis}} an expression based on a pin-hole camera model was developed to quantify the localization accuracy as a function of object size in both physical and pixel space given by: 

\vspace{-4mm}
\begin{equation}
\centering
    \begin{split}
    \Delta D= \left(\frac{f}{p_{h}} \frac{\Delta D}{D}\right)\frac{h_{LM}}{px_{LM}}
    \end{split}
\end{equation}
\vspace{-3mm}

\noindent where \(\Delta \) D is the error [m], \(D\) is the euclidean distance to the landmark [m], \(f\) is the camera focal-length [m], \(p_{h}\) is the physical height of a camera pixel [m], \(h_{LM}\) is the physical height of the landmark [m], and \(px_{LM}\) is the height of the landmark in pixel space [px]. \(\Delta D\) appears on both sides of the expression as it was found that both the results outlined here as well as in\textcolor{gray}{\cite{amzvisionthesis}} produced absolute maximum errors approximately linear with distance, implying a constant value of \(\Delta D / D\) which can be substituted into the expression. 

The bracketed term on the right hand side provides a non-dimensional constant measuring the number of pixels required to localize a landmark with given dimensions with a certain mapping accuracy. Our work results in a non-dimensional constant of 34.6 compared to 268.4 from\textcolor{gray}{\cite{amzvisionthesis}}, implying that our algorithm requires 7.7x less pixels to localize an object with the same accuracy.

To validate system latency, all 8 Xavier CPU cores were artificially stressed as each section of the pipeline was profiled. As the landmark depths from both pipelines are processed in series the latency linearly depends on the number of landmarks detected. For this study, a track was artificially set up such that the system detected 20 cones. In practice, the look-ahead results in fewer than 20 cones detected, especially in corner cases outlined in Section III where latency is most important, making these results conservative.

\begin{figure}[h]
\vspace{-4mm}
\centering
\includegraphics[width=0.45\textwidth]{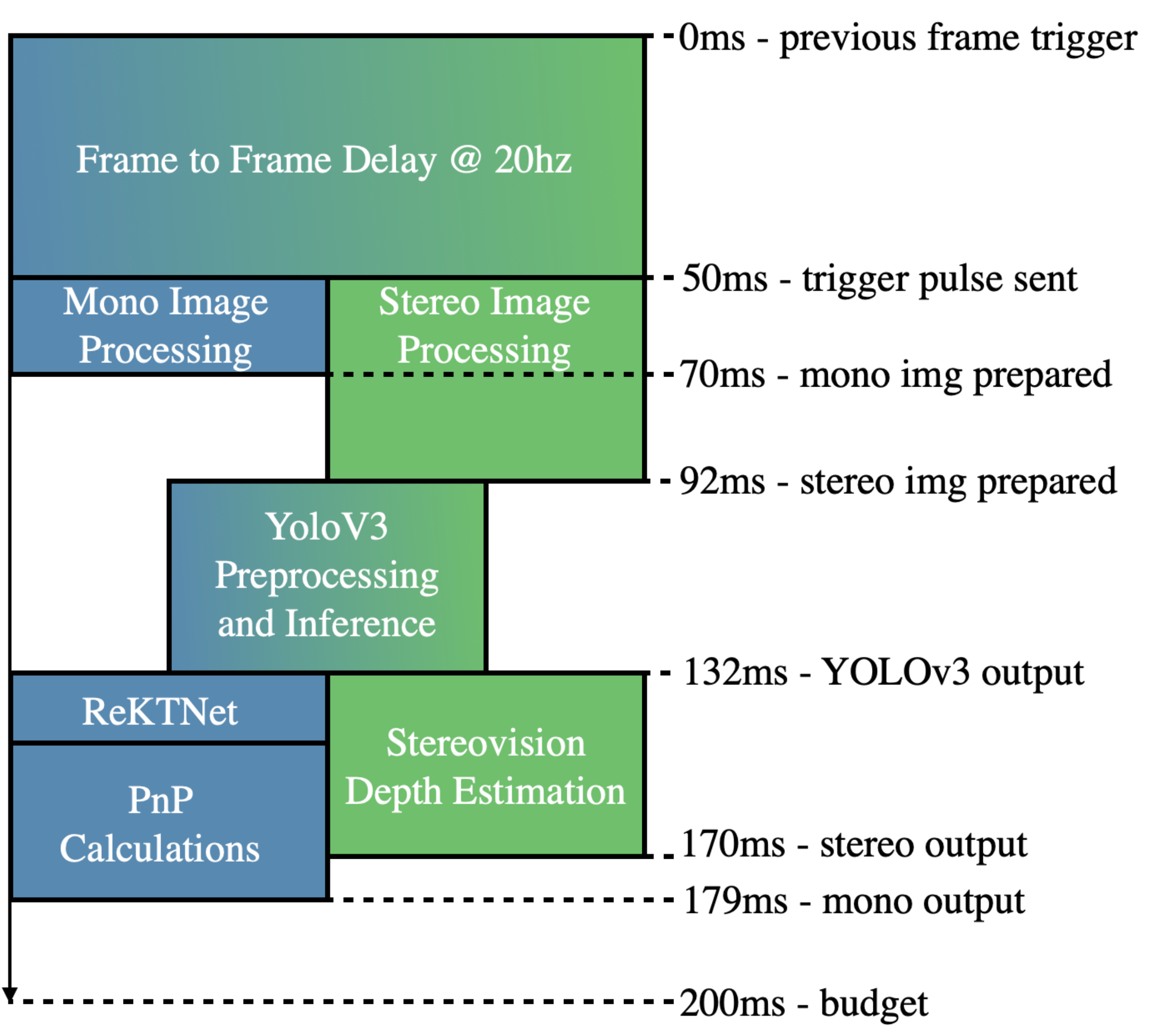}
\vspace{-4mm}
\caption{Monocular pipeline latency on the left with the stereovision pipeline on the right. The total system latency is 10\% under the 200ms requirement, including frame to frame delays which account for 28\% of the total latency.}
\vspace{-3mm}
\label{latency}
\end{figure}

As is clear from Figure~\textcolor{red}{\ref{latency}}, both pipelines running in parallel publish landmark locations within the budgeted perception system latency. This measurement includes the frame to frame delay, which accounts for scenarios where landmarks come into the FOV immediately after the camera shutter closes, and is responsible for 28\% of the total latency. This component of the latency stack is typically overlooked. To provide context for these values, the median visual perception latency of a human is 261ms for tasks far more simplistic\textcolor{gray}{\cite{human_reaction}} -- our perception system outdoes this by 45\%.

%% file: 07_Conclusions.tex
\section{CONCLUSIONS}

Accurate and low-latency visual perception is applicable to many domains, from autonomous driving to augmented reality. We present challenges that required us to optimize known solutions to develop designs for Formula Student Driverless. The end result of this work is a perception system that is able to achieve an sub-180ms latency from view to depth estimation, with errors less than 0.5m at a distance of 20m. This system and its associated source code are available for other teams as a baseline to build off of to facilitate progress in high-performance autonomy.

%% file: main.bbl
\begin{thebibliography}{10}
\providecommand{\url}[1]{#1}
\csname url@samestyle\endcsname
\providecommand{\newblock}{\relax}
\providecommand{\bibinfo}[2]{#2}
\providecommand{\BIBentrySTDinterwordspacing}{\spaceskip=0pt\relax}
\providecommand{\BIBentryALTinterwordstretchfactor}{4}
\providecommand{\BIBentryALTinterwordspacing}{\spaceskip=\fontdimen2\font plus
\BIBentryALTinterwordstretchfactor\fontdimen3\font minus
  \fontdimen4\font\relax}
\providecommand{\BIBforeignlanguage}[2]{{%
\expandafter\ifx\csname l@#1\endcsname\relax
\typeout{** WARNING: IEEEtranS.bst: No hyphenation pattern has been}%
\typeout{** loaded for the language `#1'. Using the pattern for}%
\typeout{** the default language instead.}%
\else
\language=\csname l@#1\endcsname
\fi
#2}}
\providecommand{\BIBdecl}{\relax}
\BIBdecl

\bibitem{fsg}
\emph{Autonomous Driving at Formula Student Germany 2017},
  https://www.formulastudent.de/pr/news/details/article/autonomous-driving-at-formula-student-germany-2017/.

\bibitem{fsg_results}
\emph{{Formula Student Germany Results 2019: Formula Student Driverless}},
  https://www.formulastudent.de/fsg/results/2019/.

\bibitem{fsitaly_results}
\emph{{Formula Student Italy Results}}, https://www.formula-ata.it/results-2/.

\bibitem{nerian}
\emph{Lens Focal Length and Stereo Baseline Calculator},
  https://nerian.com/support/resources/calculator/.

\bibitem{selfracingcars}
\emph{{Self Racing Cars}}, Jan 2019, http://selfracingcars.com/.

\bibitem{stereo2}
B.~Barrois, S.~Hristova, C.~Wohler, F.~Kummert, and C.~Hermes, ``{3D} pose
  estimation of vehicles using a stereo camera,'' in \emph{IEEE Intelligent
  Vehicles Symposium}, 2009.

\bibitem{deeppicar}
M.~G. Bechtel, E.~McEllhiney, and H.~Yun, ``Deeppicar: {A} low-cost deep neural
  network-based autonomous car,'' \emph{CoRR}, 2017.

\bibitem{asf}
K.~Brandes, A.~Agarwal, R.~Chang, K.~Doherty, M.~Kabir, K.~Strobel, N.~Stathas,
  C.~Trap, A.~Wang, L.~Kulik, and et~al., \emph{Robust, High Performance
  Software Design for the DUT18D Autonomous Racecar}.

\bibitem{roborace}
D.~Caporale, A.~Settimi, F.~Massa, F.~Amerotti, A.~Corti, A.~Fagiolini,
  M.~Guiggian, A.~Bicchi, and L.~Pallottino, ``Towards the design of robotic
  drivers for full-scale self-driving racing cars,'' in \emph{International
  Conference on Robotics and Automation}, 2019.

\bibitem{stereo3}
J.-R. Chang and Y.-S. Chen, ``Pyramid stereo matching network,'' in
  \emph{Proceedings of the IEEE Conference on Computer Vision and Pattern
  Recognition}, 2018.

\bibitem{joint_detection1}
L.~Chen, Z.~Yang, J.~Ma, and Z.~Luo, ``Driving scene perception network:
  Real-time joint detection, depth estimation and semantic segmentation,''
  \emph{CoRR}, 2018.

\bibitem{amzvisionthesis}
A.~Dhall, ``Real-time {3D} pose estimation with a monocular camera using deep
  learning and object priors on an autonomous racecar,'' 2018.

\bibitem{dhall2019real}
A.~Dhall, D.~Dai, and L.~Van~Gool, ``Real-time {3D} traffic cone detection for
  autonomous driving,'' in \emph{IEEE Intelligent Vehicles Symposium}, 2019.

\bibitem{nyu_depth}
D.~Eigen, C.~Puhrsch, and R.~Fergus, ``Depth map prediction from a single image
  using a multi-scale deep network,'' \emph{CoRR}, 2014.

\bibitem{etsi2013101}
T.~ETSI, ``101 539-1 v1. 1.1 (2013-08) intelligent transport systems (its),''
  \emph{V2X Applications}, 2013.

\bibitem{perceptionlatency}
D.~Falanga, S.~Kim, and D.~Scaramuzza, ``How fast is too fast? {T}he role of
  perception latency in high-speed sense and avoid,'' \emph{IEEE Robotics and
  Automation Letters}, 2019.

\bibitem{monocular3}
H.~Fu, M.~Gong, C.~Wang, K.~Batmanghelich, and D.~Tao, ``Deep ordinal
  regression network for monocular depth estimation,'' in \emph{Proceedings of
  the IEEE Conference on Computer Vision and Pattern Recognition}, 2018.

\bibitem{othercv3}
R.~Garg, V.~K. BG, G.~Carneiro, and I.~Reid, ``Unsupervised cnn for single view
  depth estimation: Geometry to the rescue,'' in \emph{European Conference on
  Computer Vision}, 2016.

\bibitem{fsg_rules}
\BIBentryALTinterwordspacing
F.~S. Germany, \emph{Formula Student Rules 2019. V1.0}. [Online]. Available:
  \url{https://www.formulastudent.de/fileadmin/user_upload/all/2019/rules/FS-Rules_2019_V1.0.pdf}
\BIBentrySTDinterwordspacing

\bibitem{monocular1}
C.~Godard, O.~Mac~Aodha, and G.~J. Brostow, ``Unsupervised monocular depth
  estimation with left-right consistency,'' in \emph{Proceedings of IEEE
  Conference on Computer Vision and Pattern Recognition}, 2017.

\bibitem{amzicra2}
N.~Gosala, A.~B{\"u}hler, M.~Prajapat, C.~Ehmke, M.~Gupta, R.~Sivanesan,
  A.~Gawel, M.~Pfeiffer, M.~B{\"u}rki, I.~Sa \emph{et~al.}, ``Redundant
  perception and state estimation for reliable autonomous racing,'' in
  \emph{International Conference on Robotics and Automation}, 2019.

\bibitem{fastslam2}
G.~Grisetti, G.~D. Tipaldi, C.~Stachniss, W.~Burgard, and D.~Nardi, ``Fast and
  accurate slam with rao--blackwellized particle filters,'' \emph{Robotics and
  Autonomous Systems}, 2007.

\bibitem{mobilenet}
A.~G. Howard, M.~Zhu, B.~Chen, D.~Kalenichenko, W.~Wang, T.~Weyand,
  M.~Andreetto, and H.~Adam, ``Mobilenets: Efficient convolutional neural
  networks for mobile vision applications,'' \emph{CoRR}, 2017.

\bibitem{amzmpc}
J.~Kabzan, L.~Hewing, A.~Liniger, and M.~N. Zeilinger, ``Learning-based model
  predictive control for autonomous racing,'' \emph{IEEE Robotics and
  Automation Letters}, 2019.

\bibitem{amzwholepaper}
J.~Kabzan, M.~d. l.~I. Valls, V.~Reijgwart, H.~F.~C. Hendrikx, C.~Ehmke,
  M.~Prajapat, A.~B{\"u}hler, N.~Gosala, M.~Gupta, R.~Sivanesan \emph{et~al.},
  ``Amz driverless: The full autonomous racing system,'' 2019.

\bibitem{stereo1}
N.~Kaempchen, U.~Franke, and R.~Ott, ``Stereo vision based pose estimation of
  parking lots using {3D} vehicle models,'' in \emph{Intelligent Vehicle
  Symposium. IEEE}, 2002.

\bibitem{Adam}
D.~P. Kingma and J.~Ba, ``Adam: A method for stochastic optimization,'' 2014.

\bibitem{fsaeteam4}
S.~Koppula, ``Learning a {CNN}-based end-to-end controller for a formula {SAE}
  racecar,'' 2017.

\bibitem{drone1}
C.~Kyrkou, G.~Plastiras, T.~Theocharides, S.~I. Venieris, and C.-S. Bouganis,
  ``Dronet: Efficient convolutional neural network detector for real-time uav
  applications,'' in \emph{Design, Automation \& Test in Europe Conference \&
  Exhibition}, 2018.

\bibitem{drone3}
J.~Lee, J.~Wang, D.~Crandall, S.~{\v{S}}abanovi{\'c}, and G.~Fox, ``Real-time,
  cloud-based object detection for unmanned aerial vehicles,'' in \emph{First
  IEEE International Conference on Robotic Computing}, 2017.

\bibitem{realtimetracking}
W.~Luo, B.~Yang, and R.~Urtasun, ``Fast and furious: Real time end-to-end {3D}
  detection, tracking and motion forecasting with a single convolutional net,''
  in \emph{Proceedings of the IEEE conference on Computer Vision and Pattern
  Recognition}, 2018.

\bibitem{drone2}
S.~Majid~Azimi, ``Shuffledet: Real-time vehicle detection network in on-board
  embedded uav imagery,'' in \emph{Proceedings of the European Conference on
  Computer Vision}, 2018.

\bibitem{fastslam1}
M.~Montemerlo, S.~Thrun, D.~Koller, B.~Wegbreit \emph{et~al.}, ``Fast{SLAM}
  2.0: An improved particle filtering algorithm for simultaneous localization
  and mapping that provably converges,'' in \emph{IJCAI}, 2003.

\bibitem{fsaeteam1}
J.~Ni and J.~Hu, ``Path following control for autonomous formula racecar:
  Autonomous formula student competition,'' in \emph{IEEE Intelligent Vehicles
  Symposium}, 2017.

\bibitem{dsnt}
A.~Nibali, Z.~He, S.~Morgan, and L.~Prendergast, ``Numerical coordinate
  regression with convolutional neural networks,'' 2018.

\bibitem{semanticsegmentation}
A.~Paszke, A.~Chaurasia, S.~Kim, and E.~Culurciello, ``Enet: {A} deep neural
  network architecture for real-time semantic segmentation,'' \emph{CoRR},
  2016.

\bibitem{yolo}
J.~Redmon and A.~Farhadi, ``Yolov3: An incremental improvement,'' 2018.

\bibitem{fasterrcnn}
S.~Ren, K.~He, R.~Girshick, and J.~Sun, ``Faster {R-CNN}: Towards real-time
  object detection with region proposal networks,'' in \emph{Advances in neural
  information processing systems}, 2015.

\bibitem{monocular4}
A.~Roy and S.~Todorovic, ``Monocular depth estimation using neural regression
  forest,'' in \emph{Proceedings of the IEEE conference on computer vision and
  pattern recognition}, 2016.

\bibitem{stereo4}
A.~Saxena, J.~Schulte, A.~Y. Ng \emph{et~al.}, ``Depth estimation using
  monocular and stereo cues.'' in \emph{IJCAI}, 2007.

\bibitem{joint_detection2}
G.~Sistu, I.~Leang, and S.~Yogamani, ``Real-time joint object detection and
  semantic segmentation network for automated driving,'' \emph{CoRR}, 2019.

\bibitem{sun2020see}
Y.~Sun, W.~Zuo, and M.~Liu, ``See the future: A semantic segmentation network
  predicting ego-vehicle trajectory with a single monocular camera,''
  \emph{IEEE Robotics and Automation Letters}, vol.~5, no.~2, pp. 3066--3073,
  2020.

\bibitem{darpa}
S.~Thrun, M.~Montemerlo, H.~Dahlkamp, D.~Stavens, A.~Aron, J.~Diebel, P.~Fong,
  J.~Gale, M.~Halpenny, G.~Hoffmann \emph{et~al.}, ``Stanley: The robot that
  won the {DARPA} grand challenge,'' \emph{Journal of field Robotics}, 2006.

\bibitem{amzicra1}
M.~I. Valls, H.~F. Hendrikx, V.~J. Reijgwart, F.~V. Meier, I.~Sa, R.~Dub{\'e},
  A.~Gawel, M.~B{\"u}rki, and R.~Siegwart, ``Design of an autonomous racecar:
  Perception, state estimation and system integration,'' in \emph{IEEE
  International Conference on Robotics and Automation}, 2018.

\bibitem{human_reaction}
D.~Woods, J.~Wyma, E.~Yund, T.~Herron, and B.~Reed, ``Factors influencing the
  latency of simple reaction time,'' \emph{Frontiers in human neuroscience},
  2015.

\bibitem{squeeze_net}
B.~Wu, F.~N. Iandola, P.~H. Jin, and K.~Keutzer, ``Squeezedet: Unified, small,
  low power fully convolutional neural networks for real-time object detection
  for autonomous driving,'' \emph{CoRR}, 2016.

\bibitem{fsaeteam3}
D.~Zadok, T.~Hirshberg, A.~Biran, K.~Radinsky, and A.~Kapoor, ``Explorations
  and lessons learned in building an autonomous formula {SAE} car from
  simulations,'' 2019.

\bibitem{fsaeteam2}
M.~Zeilinger, R.~Hauk, M.~Bader, and A.~Hofmann, ``Design of an autonomous race
  car for the formula student driverless,'' 2017.

\bibitem{monocular2}
H.~Zhan, R.~Garg, C.~Saroj~Weerasekera, K.~Li, H.~Agarwal, and I.~Reid,
  ``Unsupervised learning of monocular depth estimation and visual odometry
  with deep feature reconstruction,'' in \emph{Proceedings of the IEEE
  Conference on Computer Vision and Pattern Recognition}, 2018.

\bibitem{othercv2}
L.~Zhang, L.~Lin, X.~Liang, and K.~He, ``Is faster {R-CNN} doing well for
  pedestrian detection?'' in \emph{European conference on computer vision},
  2016.

\bibitem{othercv1}
X.~Zhou, M.~Zhu, S.~Leonardos, K.~G. Derpanis, and K.~Daniilidis, ``Sparseness
  meets deepness: {3D} human pose estimation from monocular video,'' in
  \emph{Proceedings of the IEEE conference on computer vision and pattern
  recognition}, 2016.

\end{thebibliography}
